# Symbolic Dynamic Programming for Discrete and Continuous State MDPs

**Scott Sanner**
NICTA & the ANU
Canberra, Australia
ssanner@nicta.com.au

**Karina Valdivia Delgado**
University of Sao Paulo
Sao Paulo, Brazil
kvd@ime.usp.br

**Leliane Nunes de Barros**
University of Sao Paulo
Sao Paulo, Brazil
leliane@ime.usp.br

## Abstract

Many real-world decision-theoretic planning problems can be naturally modeled with discrete and continuous state Markov decision processes (DC-MDPs). While previous work has addressed automated decision-theoretic planning for DC-MDPs, *optimal* solutions have only been defined so far for limited settings, e.g., DC-MDPs having *hyper-rectangular piecewise linear value functions*. In this work, we extend symbolic dynamic programming (SDP) techniques to provide optimal solutions for a vastly expanded class of DC-MDPs. To address the inherent combinatorial aspects of SDP, we introduce the XADD — a continuous variable extension of the algebraic decision diagram (ADD) — that maintains compact representations of the exact value function. Empirically, we demonstrate an implementation of SDP with XADDs on various DC-MDPs, showing the *first optimal automated solutions* to DC-MDPs with *linear and nonlinear piecewise partitioned value functions* and showing the advantages of constraint-based pruning for XADDs.

## 1 Introduction

Many real-world stochastic planning problems involving resources, time, or spatial configurations naturally use continuous variables in their state representation. For example, in the MARS ROVER problem [6], a rover must manage bounded continuous resources of battery power and daylight time as it plans scientific discovery tasks for a set of landmarks on a given day.

While problems such as the MARS ROVER are naturally modeled by discrete and continuous state Markov decision processes (DC-MDPs), little progress seems to have been made in recent years in developing *exact* solutions for DC-MDPs with multiple continuous state variables beyond the subset of DC-MDPs which have an optimal *hyper-rectangular piecewise linear value function* [8, 11].

Yet even simple DC-MDPs may require optimal value functions that are piecewise functions with non-rectangular boundaries; as an illustration, we consider KNAPSACK:

**Example 1.1** (KNAPSACK)**.** *We have three continuous state variables: $k \in [0, 100]$ indicating knapsack weight, and two sources of knapsack contents: $x_i \in [0, 100]$ for $i \in \{1, 2\}$. We have two actions $move_i$ for $i \in \{1, 2\}$ that can move* **all** *of a resource from $x_i$ to the knapsack* **if** *the knapsack weight remains below its capacity of* 100*. We get an immediate reward for any weight added to the knapsack.*

*We can formalize the transition and reward for* KNAPSACK *action $move_i$ ($i \in \{1, 2\}$) using conditional equations, where $(k, x_1, x_2)$ and $(k', x'_1, x'_2)$ are respectively the pre- and post-action state and $R$ is immediate reward:*

$$k' = \begin{cases} k + x_i \leq 100: & k + x_i \\ k + x_i > 100: & k \end{cases}$$

$$R = \begin{cases} k + x_i \leq 100: & x_i \\ k + x_i > 100: & 0 \end{cases}$$

$$x'_i = \begin{cases} k + x_i \leq 100: & 0 \\ k + x_i > 100: & x_i \end{cases}$$

$$x'_j = x_j, \ (j \neq i)$$

If our objective is to maximize the long-term *value* $V$ (i.e., the sum of rewards received over an infinite horizon of actions), then we can write the optimal value achievable from a given state in KNAPSACK as a function of state variables:

$$V = \begin{cases} x_1 + k > 100 \wedge x_2 + k > 100: & 0 \\ x_1 + k > 100 \wedge x_2 + k \leq 100: & x_2 \\ x_1 + k \leq 100 \wedge x_2 + k > 100: & x_1 \\ x_1 + k \leq 100 \wedge x_2 + k \leq 100 \wedge x_2 > x_1: & x_2 \\ x_1 + k \leq 100 \wedge x_2 + k \leq 100 \wedge x_2 \leq x_1: & x_1 \\ x_1 + x_2 + k \leq 100: & x_1 + x_2 \end{cases} \quad (1)$$

One will see that this encodes the following rules (in order): (a) if both resources are too large for the knapsack, 0 reward is obtained, (b) otherwise if only one item can fit, the reward is for the largest item that fits, (c) otherwise if both items can fit then reward $x_1 + x_2$ is obtained. Here we note that the value function is piecewise linear, but it contains decision boundaries like $x_1 + x_2 + k \leq 100$ that are clearly non-rectangular; rectangular boundaries are restricted to conjunctions of simple inequalities of a continuous variable and a constant (e.g., $x_1 \leq 5 \wedge x_2 > 2 \wedge k \geq 0$).

What is interesting to note is that although KNAPSACK is very simple, no previous algorithm in the DC-MDP literature has been proposed to exactly solve it due to the nature of its non-rectangular piecewise optimal value function. Of course our focus in this paper is not just on KNAPSACK — researchers have spent decades finding improved solutions to this particular combinatorial optimization problem — but rather on general stochastic sequential optimization in DC-MDPs that contain structure similar to KNAPSACK, as well as highly nonlinear structure beyond KNAPSACK. Both types of problem structure are exemplified in the MARS ROVER problems we experiment on later.

In proposing a solution to these problems, an important question arises: if the solution to KNAPSACK is simple and intuitive, why is it beyond the reach of existing exact DC-MDP solutions? In response, it seems that it has not been clear what value function representation supports closed-form computation of the Bellman backup (regression and maximization operations) for general DC-MDP transition and reward structures. These questions have been affirmatively addressed for the subset of DC-MDPs with transition functions that are mixtures of delta functions and reward functions that are hyper-rectangular piecewise linear, which provably lead to value functions of the same structure [8, 11]. However, the literature appears to lack a solution to this problem when, for example, the reward instead uses piecewise nonlinear functions with linear or nonlinear boundaries, leading to value functions of similar structure.

In this paper, we propose novel ideas to workaround some of the expressiveness limitations of previous approaches and significantly generalize the range of DC-MDPs that can be solved exactly. To achieve this more general solution, this paper contributes a number of important advances:

- We propose to represent the transition function of a DC-MDP using conditional stochastic equations; in using this formalism, we observe that many aspects of the proposed symbolic DC-MDP solution become readily apparent.

- The use of conditional stochastic equations facilitates symbolic regression of the value function via substitutions. This is precisely the motivation behind symbolic dynamic programming (SDP) [4] used to solve MDPs with transitions and reward functions defined in first-order logic, except that in prior SDP work, only piecewise constant functions have been used; in this work we introduce techniques for working with *arbitrary* piecewise symbolic functions.

- While the *case* representation for the optimal KNAPSACK solution shown in (1) is sufficient in theory to represent the optimal value functions that our DC-MDP solution produces, this representation is unreasonable to maintain in practice since the number of case partitions may grow exponentially on each receding horizon control step. For *discrete* factored MDPs, algebraic decision diagrams (ADDs) [1] have been successfully used in exact algorithms like SPUDD [9] to maintain compact value representations. Motivated by this work we introduce extended ADDs (XADDs) to compactly represent general piecewise functions and show how to perform efficient operations on them *including* symbolic maximization. We also borrow techniques from [14] for constraint-based pruning of XADDs that can be applied when XADDs meet certain expressiveness restrictions.

Aided by these algorithmic and data structure advances, we empirically demonstrate that our SDP approach with XADDs can exactly solve a variety of DC-MDPs with *general piecewise linear and nonlinear value functions* for which no previous analytical solution has been proposed.

## 2 Discrete and Continuous State MDPs

We first introduce discrete and continuous state Markov decision processes (DC-MDPs) and then review their finite-horizon solution via dynamic programming following [11].

### 2.1 Factored Representation

In a DC-MDP, states will be represented by vectors of variables $(\vec{b}, \vec{x}) = (b_1, \ldots, b_n, x_1, \ldots, x_m)$. We assume that each state variable $b_i$ $(1 \leq i \leq n)$ is boolean s.t. $b_i \in \{0, 1\}$ and each $x_j$ $(1 \leq j \leq m)$ is continuous s.t. $x_j \in [L_j, U_j]$ for $L_j, U_j \in \mathbb{R}; L_j \leq U_j$. We also assume a finite set of actions $A = \{a_1, \ldots, a_p\}$.

A DC-MDP is defined by the following: (1) a state transition model $P(\vec{b}', \vec{x}' | \cdots, a)$, which specifies the probability of the next state $(\vec{b}', \vec{x}')$ conditioned on a subset of the previous and next state (defined below) and action $a$; (2) a reward function $R(\vec{b}, \vec{x}, a)$, which specifies the immediate reward obtained by taking action $a$ in state $(\vec{b}, \vec{x})$; and (3) a discount factor $\gamma$, $0 \leq \gamma \leq 1$.[1] A policy $\pi$ specifies the action $\pi(\vec{b}, \vec{x})$ to take in each state $(\vec{b}, \vec{x})$. Our goal is to find an optimal sequence of horizon-dependent policies $\Pi^* = (\pi^{*,1}, \ldots, \pi^{*,H})$ that maximizes the expected sum

[1] If time is explicitly included as one of the continuous state variables, $\gamma = 1$ is typically used, unless discounting by horizon (different from the state variable time) is still intended.

of discounted rewards over a horizon $h \in H; H \geq 0$:[2]

$$V^{\Pi^*}(\vec{x}) = E_{\pi^*}\left[\sum_{h=0}^{H} \gamma^h \cdot r^h \middle| \vec{b}_0, \vec{x}_0\right], \qquad (2)$$

Here $r^h$ is the reward obtained at horizon $h$ following $\Pi^*$ where we assume starting state $(\vec{b}_0, \vec{x}_0)$ at $h = 0$.

DC-MDPs as defined above are naturally factored [3] in terms of state variables $(\vec{b}, \vec{x})$; as such transition structure can be exploited in the form of a dynamic Bayes net (DBN) [7] where the individual conditional probabilities $P(b_i'|\cdots, a)$ and $P(x_j'|\cdots, a)$ condition on a subset of the variables in the current and next state. We disallow *synchronic arcs* (variables that condition on each other in the same time slice) within the binary $\vec{b}$ and continuous variables $\vec{x}$, but we allow synchronic arcs from $\vec{b}$ to $\vec{x}$ (note that these conditions enforce the directed acyclic graph requirements of DBNs). Thus, the joint transition model can be specified as

$$P(\vec{b}', \vec{x}'|\cdots, a) = \qquad (3)$$
$$\prod_{i=1}^{n} P(b_i'|\vec{b}, \vec{x}, a) \prod_{j=1}^{m} P(x_j'|\vec{b}, \vec{b}', \vec{x}, a)$$

where $P(b_i'|\vec{b}, \vec{x}, a)$ may condition on a subset of $\vec{b}$ and $\vec{x}$ and likewise $P(x_j'|\vec{b}, \vec{b}', \vec{x}, a)$ may condition on a subset of $\vec{b}$, $\vec{b}'$, and $\vec{x}$.

As for standard finite discrete factored MDPs, the conditional probabilities $P(b_i'|\vec{b}, \vec{x}, a)$ for *binary* variables $b_i$ $(1 \leq i \leq n)$ can be represented by conditional probability tables (CPTs). For the *continuous* variables $x_j$ $(1 \leq j \leq m)$, we represent the continuous probability functions (CPFs) $P(x_j'|\vec{b}, \vec{b}', \vec{x}, a)$ with *conditional stochastic equations* (CSEs). For the solution provided here, we only require two properties of these CSEs: (1) they are *Markov*, meaning that they can only condition on the previous state, and (2) they are *deterministic* meaning that the next state must be uniquely determined from the previous state (i.e., $x_1' = x_1 + x_2^2$ is deterministic whereas $x_1'^2 = x_1^2$ is not because $x_1' = \pm x_1$).[3] Otherwise, we allow for arbitrary functions in these Markovian, conditional deterministic equations as in the following example:

$$P(x_1'|\vec{b}, \vec{b}', \vec{x}, a) = \delta\left[x_1' - \begin{cases} b_1' \wedge x_2^2 \leq 1: & \exp(x_1^2 - x_2^2) \\ \neg b_1' \vee x_2^2 > 1: & x_1 + x_2 \end{cases}\right] \qquad (4)$$

Here the use of the Dirac $\delta[\cdot]$ function ensures that this is a conditional probability function that integrates to 1 over $x_1'$ in this case. But in more intuitive terms, one can see that this $\delta[\cdot]$ encodes the deterministic transition equation $x_1' = \ldots$ where $\ldots$ is the conditional portion of (4). In this work, we require all CSEs in the transition function for variable $x_i'$ to use the $\delta[\cdot]$ as shown in this example.

It will be obvious that CSEs in the form of (4) are *conditional equations*; they are furthermore *stochastic* because they can condition on boolean random variables in the same time slice that are stochastically sampled, e.g., $b_1'$ in (4). Of course, these CSEs are restricted in that they cannot represent general stochastic noise (e.g., Gaussian noise), but we note that this representation effectively allows modeling of continuous variable transitions as a mixture of $\delta$ functions, which has been used heavily in previous exact DC-MDP solutions [8, 11, 13]. Furthermore, we note that our representation is more general than [8, 11, 13] in that we do not restrict the equation to be linear, but rather allow it to specify *arbitrary* functions (e.g., nonlinear) as demonstrated in (4).

We allow the reward function $R_a(\vec{b}, \vec{x})$ to be *any* arbitrary function of the current state for each action $a \in A$, for example:

$$R_a(\vec{b}, \vec{x}) = \begin{cases} x_1^2 + x_2^2 \leq 1: & 1 - x_1^2 - x_2^2 \\ x_1^2 + x_2^2 > 1: & 0 \end{cases} \qquad (5)$$

or even

$$R_a(\vec{b}, \vec{x}) = 10 x_3 x_4 \exp(x_1^2 + \sqrt{x_2}) \qquad (6)$$

While our DC-MDP examples throughout the paper will demonstrate the full expressiveness of our symbolic dynamic programming approach, we note that there are computational advantages to be had when the reward and transition case conditions and functions can be restricted, e.g., to polynomials. We will return to this issue later.

## 2.2 Solution Methods

Now we provide a continuous state generalization of *value iteration* [2], which is a dynamic programming algorithm for constructing optimal policies. It proceeds by constructing a series of $h$-stage-to-go value functions $V^h(\vec{b}, \vec{x})$. Initializing $V^0(\vec{b}, \vec{x})$ (e.g., to $V^0(\vec{b}, \vec{x}) = 0$) we define the quality of taking action $a$ in state $(\vec{b}, \vec{x})$ and acting so as to obtain $V^h(\vec{b}, \vec{x})$ thereafter as the following:

$$Q_a^{h+1}(\vec{b}, \vec{x}) = R_a(\vec{b}, \vec{x}) + \gamma \cdot \qquad (7)$$
$$\sum_{\vec{b}'} \int_{\vec{x}'} \left(\prod_{i=1}^{n} P(b_i'|\vec{b}, \vec{x}, a) \prod_{j=1}^{m} P(x_j'|\vec{b}, \vec{b}', \vec{x}, a)\right) V^h(\vec{b}', \vec{x}') d\vec{x}'$$

Given $Q_a^h(\vec{b}, \vec{x})$ for each $a \in A$, we can proceed to define the $h+1$-stage-to-go value function as follows:

$$V^{h+1}(\vec{b}, \vec{x}) = \max_{a \in A}\left\{Q_a^{h+1}(\vec{b}, \vec{x})\right\} \qquad (8)$$

[2] $H = \infty$ is allowed if an optimal policy has a finitely bounded value (guaranteed if $\gamma < 1$); for $H = \infty$, the optimal policy is independent of horizon, i.e., $\forall h \geq 0, \pi^{*,h} = \pi^{*,h+1}$.

[3] While the *deterministic* requirement may seem to conflict with the label of *stochastic*, we note that stochasticity enters through the conditional component, to be discussed in a moment.

If the horizon $H$ is finite, then the optimal value function is obtained by computing $V^H(\vec{b},\vec{x})$ and the optimal horizon-dependent policy $\pi^{*,h}$ at each stage $h$ can be easily determined via $\pi^{*,h}(\vec{b},\vec{x}) = \arg\max_a Q_a^h(\vec{b},\vec{x})$. If the horizon $H = \infty$ and the optimal policy has finitely bounded value, then value iteration can terminate at horizon $h+1$ if $V^{h+1} = V^h$; then $\pi^*(\vec{b},\vec{x}) = \arg\max_a Q_a^{h+1}(\vec{b},\vec{x})$.

Of course this is simply the *mathematical* definition. In the discrete-only case, we can always compute this in tabular form; however, how to compute this for DC-MDPs with reward and transition function as previously defined is the objective of the symbolic dynamic programming algorithm that we define next.

## 3 Symbolic Dynamic Programming

As it's name suggests, symbolic dynamic programming (SDP) [4] is simply the process of performing dynamic programming (in this case value iteration) via symbolic manipulation. While SDP as defined in [4] was previously only used with piecewise constant functions, we now generalize the representation to work with general piecewise functions needed for DC-MDPs in this paper.

Before we define our solution, however, we must formally define our case representation and symbolic case operators.

### 3.1 Case Representation and Operators

Throughout this paper, we will assume that all symbolic functions can be represented in *case* form as follows:

$$f = \begin{cases} \phi_1 & f_1 \\ \vdots & \vdots \\ \phi_k & f_k \end{cases}$$

Here the $\phi_i$ are logical formulae defined over the state $(\vec{b},\vec{x})$ that can include arbitrary logical $(\wedge, \vee, \neg)$ combinations of (a) boolean variables in $\vec{b}$ and (b) inequalities $(\geq, >, \leq, <)$, equalities $(=)$, or disequalities $(\neq)$ where the left and right operands can be *any* function of one or more variables in $\vec{x}$. Each $\phi_i$ will be disjoint from the other $\phi_j$ $(j \neq i)$; however the $\phi_i$ may not exhaustively cover the state space, hence $f$ may only be a *partial function* and may be undefined for some state assignments. The $f_i$ can be *any* functions of the state variables in $\vec{x}$.

As concrete examples, consider the transition representation for KNAPSACK in Ex. 1.1, the optimal value function for KNAPSACK from (1), or any of (4), (5), or (6).

*Unary operations* such as scalar multiplication $c \cdot f$ (for some constant $c \in \mathbb{R}$) or negation $-f$ on case statements $f$ are straightforward; the unary operation is simply applied to each $f_i$ $(1 \leq i \leq k)$. Intuitively, to perform a *binary operation* on two case statements, we simply take the cross-product of the logical partitions of each case statement and perform the corresponding operation on the resulting paired partitions. Letting each $\phi_i$ and $\psi_j$ denote generic first-order formulae, we can perform the "cross-sum" $\oplus$ of two (unnamed) cases in the following manner:

$$\begin{cases} \phi_1: & f_1 \\ \phi_2: & f_2 \end{cases} \oplus \begin{cases} \psi_1: & g_1 \\ \psi_2: & g_2 \end{cases} = \begin{cases} \phi_1 \wedge \psi_1: & f_1 + g_1 \\ \phi_1 \wedge \psi_2: & f_1 + g_2 \\ \phi_2 \wedge \psi_1: & f_2 + g_1 \\ \phi_2 \wedge \psi_2: & f_2 + g_2 \end{cases}$$

Likewise, we can perform $\ominus$ and $\otimes$ by, respectively, subtracting or multiplying partition values (as opposed to adding them) to obtain the result. Some partitions resulting from the application of the $\oplus$, $\ominus$, and $\otimes$ operators may be inconsistent (infeasible); we may simply discard such partitions as they are irrelevant to the function value.

For SDP, we'll also need to perform maximization, restriction, and substitution on case statements. *Symbolic maximization* is fairly straightforward to define:

$$\max\left( \begin{cases} \phi_1: & f_1 \\ \phi_2: & f_2 \end{cases}, \begin{cases} \psi_1: & g_1 \\ \psi_2: & g_2 \end{cases} \right) = \begin{cases} \phi_1 \wedge \psi_1 \wedge f_1 > g_1: & f_1 \\ \phi_1 \wedge \psi_1 \wedge f_1 \leq g_1: & g_1 \\ \phi_1 \wedge \psi_2 \wedge f_1 > g_2: & f_1 \\ \phi_1 \wedge \psi_2 \wedge f_1 \leq g_2: & g_2 \\ \phi_2 \wedge \psi_1 \wedge f_2 > g_1: & f_2 \\ \phi_2 \wedge \psi_1 \wedge f_2 \leq g_1: & g_1 \\ \phi_2 \wedge \psi_2 \wedge f_2 > g_2: & f_2 \\ \phi_2 \wedge \psi_2 \wedge f_2 \leq g_2: & g_2 \end{cases}$$

One can verify that the resulting case statement is still within the case language defined previously. At first glance this may seem like a cheat and little is gained by this symbolic sleight of hand. However, simply having a case partition representation that is closed under maximization will facilitate the closed-form regression step that we need for SDP. Furthermore, the XADD that we introduce later will be able to exploit the internal decision structure of this maximization to represent it much more compactly.

The next operation of *restriction* is fairly simple: in this operation, we want to restrict a function $f$ to apply only in cases that satisfy some formula $\phi$, which we write as $f|_\phi$. This can be done by simply appending $\phi$ to each case partition as follows:

$$f = \begin{cases} \phi_1: & f_1 \\ \vdots & \vdots \\ \phi_k: & f_k \end{cases} \qquad f|_\phi = \begin{cases} \phi_1 \wedge \phi: & f_1 \\ \vdots & \vdots \\ \phi_k \wedge \phi: & f_k \end{cases}$$

Clearly $f|_\phi$ only applies when $\phi$ holds and is undefined otherwise, hence $f|_\phi$ is a partial function unless $\phi \equiv \top$.

The final operation that we need to define for case statements is substitution. *Symbolic substitution* simply takes a set $\sigma$ of variables and their substitutions, e.g., $\sigma = \{x_1' = x_1 + x_2, x_2' = x_1^2 \exp(x_2)\}$ where the LHS of the $=$ represents the substitution variable and the RHS of the $=$ represents the expression that should be substituted in its place.

No variable occurring in any RHS expression of $\sigma$ can also occur in any LHS expression of $\sigma$. We write the substitution of a non-case function $f_i$ with $\sigma$ as $f_i\sigma$; as an example, for the $\sigma$ defined previously and $f_i = x'_1 + x'_2$ then $f_i\sigma = x_1 + x_2 + x_1^2 \exp(x_2)$ as would be expected. We can also substitute into case partitions $\phi_j$ by applying $\sigma$ to its LHS and RHS operands; as an example, if $\phi_j \equiv x'_1 \leq \exp(x'_2)$ then $\phi_j\sigma \equiv x_1 + x_2 \leq \exp(x_1^2 \exp(x_2))$. Having now defined substitution of $\sigma$ for non-case functions $f_i$ and case partitions $\phi_j$ we can define it for case statements in general:

$$f = \begin{cases} \phi_1 : & f_1 \\ \vdots & \vdots \\ \phi_k : & f_k \end{cases} \qquad f\sigma = \begin{cases} \phi_1\sigma : & f_1\sigma \\ \vdots & \vdots \\ \phi_k\sigma : & f_k\sigma \end{cases}$$

One property of substitution is that if $f$ has mutually exclusive partitions $\phi_i$ ($1 \leq i \leq k$) then $f\sigma$ must also have mutually exclusive partitions — this follows from the logical consequence that if $\phi_1 \wedge \phi_2 \models \perp$ then $\phi_1\sigma \wedge \phi_2\sigma \models \perp$.

### 3.2 Symbolic Dynamic Programming (SDP)

In the SDP solution for DC-MDPs, our objective will be to take a DC-MDP as defined in Section 2, apply value iteration as defined in Section 2.2, and produce the final value optimal function $V^h$ at horizon $h$ in the form of a case statement.

For the base case of $h = 0$, we note that setting $V^0(\vec{b}, \vec{x}) = 0$ (or to the reward case statement, if not action dependent) is trivially in the form of a case statement.

Next, $h > 0$ requires the application of SDP. Fortunately, given our previously defined operations, SDP is straightforward and can be divided into four steps:

1. *Prime the Value Function*: Since $V^h$ will become the "next state" in value iteration, we setup a substitution $\sigma = \{b_1 = b'_1, \ldots, b_n = b'_n, x_1 = x'_1, \ldots, x_m = x'_m\}$ and obtain $V'^h = V^h\sigma$.

2. *Continuous Integration*: Now that we have our primed value function $V'^h$ in case statement format defined over next state variables $(\vec{b}', \vec{x}')$, we first evaluate the integral marginalization $\int_{\vec{x}'}$ over the continuous variables in (7). Because the lower and upper integration bounds are respectively $-\infty$ and $\infty$ and we have disallowed synchronic arcs between variables in $\vec{x}'$ in the transition DBN, we can marginalize out each $x'_j$ independently, and in any order. Using *variable elimination* [17], when marginalizing over $x'_j$ we can factor out any functions independent of $x'_j$ — that is, for $\int_{x'_j}$ in (7), one can see that initially, the only functions that can include $x'_j$ are $V'^h$ and $P(x'_j|\vec{b}, \vec{b}', \vec{x}, a) = \delta[x'_j - g(\vec{x})]$; hence, the first marginal over $x'_j$ need only be computed over $\delta[x'_j - g(\vec{x})]V'^h$.

   What follows is one of the *key novel insights of SDP* in the context of DC-MDPs — the integration $\int_{x'_j} \delta[x'_j - g(\vec{x})]V'^h dx'_j$ simply *triggers the substitution* $\sigma = \{x'_j = g(\vec{x})\}$ on $V'^h$, that is

   $$\int_{x'_j} \delta[x'_j - g(\vec{x})]V'^h dx'_j \;=\; V'^h\{x'_j = g(\vec{x})\}. \quad (9)$$

   Thus we can perform the operation in (9) repeatedly in sequence *for each* $x'_j$ ($1 \leq j \leq m$) for every action $a$. The only additional complication is that the form of $P(x'_j|\vec{b}, \vec{x}, a)$ is a *conditional* equation, c.f. (4), and represented generically as follows:

   $$P(x'_j|\vec{b}, \vec{x}, a) = \delta \left[ x'_j = \begin{cases} \phi_1 : & f_1 \\ \vdots & \vdots \\ \phi_k : & f_k \end{cases} \right] \quad (10)$$

   Hence to perform (9) on this more general representation, we obtain that $\int_{x'_j} P(x'_j|\vec{b}, \vec{x}, a)V'^h dx'_j$

   $$= \begin{cases} \phi_1 : & V'^h\{x'_j = f_1\} \\ \vdots & \vdots \\ \phi_k : & V'^h\{x'_j = f_k\} \end{cases}$$

   In effect, we can read (10) as a *conditional substitution*, i.e., in each of the different *previous state* conditions $\phi_i$ ($1 \leq i \leq k$), we obtain a different substitution for $x'_j$ appearing in $V'^h$ (i.e., $\sigma = \{x'_j = f_i\}$). Here we note that because $V'^h$ is *already* a case statement, we simply replace the single partition $\phi_i$ with the multiple partitions of $V\{x'_j = f_i\}|_{\phi_i}$. This reduces the *nested* case statement back down to a non-nested case statement as in the following example:

   $$\begin{cases} \phi_1 : & \begin{cases} \psi_1 : & f_{11} \\ \psi_2 : & f_{12} \end{cases} \\ \phi_2 : & \begin{cases} \psi_1 : & f_{21} \\ \psi_2 : & f_{22} \end{cases} \end{cases} = \begin{cases} \phi_1 \wedge \psi_1 : & f_{11} \\ \phi_1 \wedge \psi_2 : & f_{12} \\ \phi_2 \wedge \psi_1 : & f_{21} \\ \phi_2 \wedge \psi_2 : & f_{22} \end{cases}$$

   To perform the full continuous integration, if we initialize $\tilde{Q}_a^{h+1} := V'^h$ for each action $a \in A$, and repeat the above integrals for all $x'_j$, updating $\tilde{Q}_a^{h+1}$ each time, then after elimination of all $x'_j$ ($1 \leq j \leq m$), we will have the partial regression of $V'^h$ for the continuous variables for each action $a$ denoted by $\tilde{Q}_a^{h+1}$.

3. *Discrete Marginalization*: Now that we have our partial regression $\tilde{Q}_a^{h+1}$ for each action $a$, we proceed to derive the full backup $Q_a^{h+1}$ from $\tilde{Q}_a^{h+1}$ by evaluating the discrete marginalization $\sum_{\vec{b}'}$ in (7). Because we previously disallowed synchronic arcs between the variables in $\vec{b}'$ in the transition DBN, we can sum out each variable $b'_i$ ($1 \le i \le n$) independently. Hence, initializing $Q_a^{h+1} := \tilde{Q}_a^{h+1}$ we perform the discrete regression by applying the following iterative process *for each* $b'_i$ in any order for each action $a$:

$$Q_a^{h+1} := \left[ Q_a^{h+1} \otimes P(b'_i|\vec{b}, \vec{x}, a) \right] |_{b'_i = \top} \oplus \left[ Q_a^{h+1} \otimes P(b'_i|\vec{b}, \vec{x}, a) \right] |_{b'_i = \bot}. \quad (11)$$

This requires a variant of the earlier restriction operator $|_v$ that actually *sets* the variable $v$ to the given value if present. Note that both $Q_a^{h+1}$ and $P(b'_i|\vec{b}, \vec{x}, a)$ can be represented as case statements (discrete CPTs *are* case statements), and each operation produces a case statement. Thus, once this process is complete, we have marginalized over all $\vec{b}'$ and $Q_a^{h+1}$ is the symbolic representation of the intended Q-function.

4. *Maximization*: Now that we have $Q_a^{h+1}$ in case format for each action $a \in \{a_1, \ldots, a_p\}$, obtaining $V^{h+1}$ in case format as defined in (8) requires sequentially applying *symbolic maximization* as defined previously:

$$V^{h+1} = \max(Q_{a_1}^{h+1}, \max(\ldots, \max(Q_{a_{p-1}}^{h+1}, Q_{a_p}^{h+1})))$$

By induction, because $V^0$ is a case statement and applying SDP to $V^h$ in case statement form produces $V^{h+1}$ in case statement form, we have achieved our intended objective with SDP. On the issue of correctness, we note that each operation above simply implements one of the dynamic programming operations in (7) or (8), so correctness simply follows from verifying (a) that each case operation produces the correct result and that (b) each case operation is applied in the correct sequence as defined in (7) or (8).

On a final note, we observe that SDP holds for *any* symbolic case statements; we have not restricted ourselves to rectangular piecewise functions, piecewise linear functions, or even piecewise polynomial functions. As the SDP solution is purely symbolic, SDP applies to *any* DC-MDPs using bounded symbolic function that can be written in case format! Of course, that is the theory, next we meet practice.

## 4 Extended ADDs (XADDs)

In practice, it can be prohibitively expensive to maintain a case statement representation of a value function with explicit partitions. Motivated by the SPUDD [9] algorithm which maintains compact value function representations for finite discrete factored MDPs using algebraic decision diagrams (ADDs) [1], we extend this formalism to handle continuous variables in a data structure we refer to as the XADD. An example XADD for the optimal KNAPSACKvalue function from (1) is provided in Figure 1.

In brief we note that an XADD is like an ADD except that (a) the decision nodes can have arbitrary inequalities, equalities, or disequalities (one per node) and (b) the leaf nodes can represent arbitrary functions. The decision nodes still have a fixed order from root to leaf and the standard ADD operations to build a canonical ADD (REDUCE) and to perform a binary operation on two ADDs (APPLY) still applies in the case of XADDs.

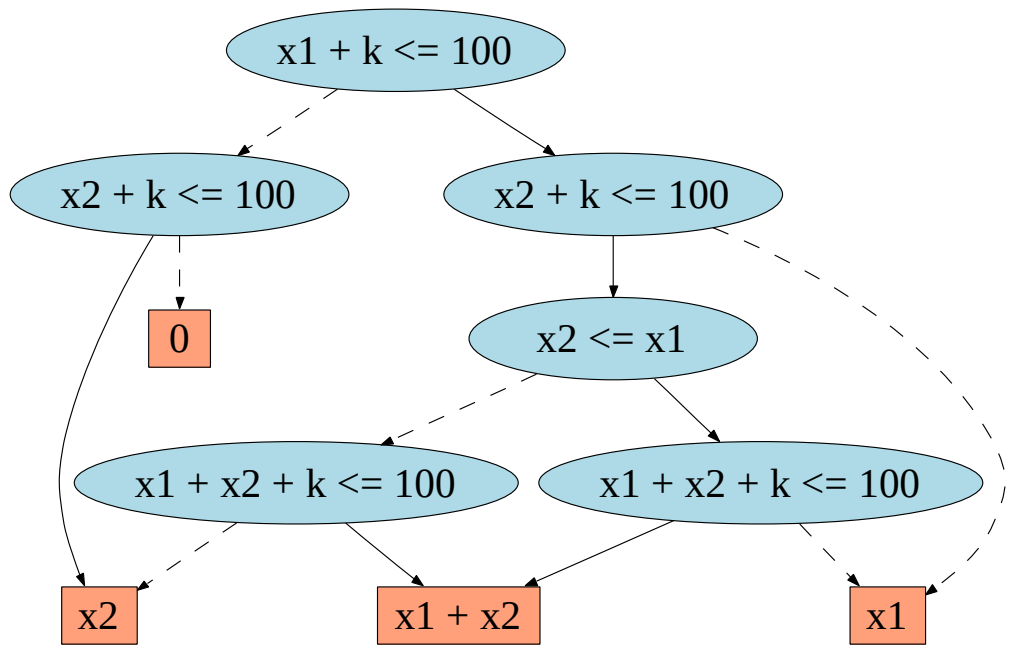

Figure 1: The optimal value function for KNAPSACK as a decision diagram: the *true* branch is solid, the *false* branch is dashed.

While exact solutions using symbolic dynamic programming are possible in principle for arbitrary symbolic CSE transition and reward functions, we note that it is much more difficult to devise a canonical and compact form for representations such as (6) in comparison to (5). Hence while we have used general examples throughout the paper to demonstrate the expressiveness of our approach, we will restrict XADDs to use *polynomial* functions only. We note the main advantage of this for the XADD is that we can put the leaf and decision nodes in a *unique, canonical* form, which allows us to minimize redundancy in the XADD representation of a case statement.

It is fairly straightforward for XADDs to support all case operations required for SDP. Standard operations like unary multiplication, negation, $\oplus$, and $\otimes$ are implemented exactly as they are for ADDs. The fact that the decision nodes have internal structure is irrelevant, although this means that certain paths in the XADD may be inconsistent or infeasible (due to parent decisions). To remedy this, when the XADD has only linear decision nodes, we can use the feasibility checkers of a linear programming solver (e.g., as also done in [14]) to prune unreachable nodes in the XADD; later we show results demonstrating impressive reductions in XADD size using this style of pruning.

The only two XADD operations that pose difficulty are substitution and maximization. In principle substitution is simple, the only caveat is that substitutions modify the decision nodes and hence decision nodes may become unordered. We can use the recursive application of ADD binary operations $\otimes$ and $\oplus$ as given in Algorithm 1 to correctly reorder the nodes in an XADD $F$ after substitution. A related reordering issue occurs during XADD maximization; because XADD maximization can introduce new de-

**Algorithm 1**: REORDER(F)

**input** : $F$ (root node for possibly unordered XADD)
**output**: $F_r$ (root node for an ordered XADD)

**begin**
  //if terminal node, return canonical terminal node
  **if** *$F$ is terminal node* **then**
    **return** *canonical terminal node for polynomial of $F$*;
  //nodes have a *true* & *false* branch and *var* id
  **if** *$F \rightarrow F_r$ is not in Cache* **then**
    $F_{true}$ = REORDER ($F_{true}$) $\otimes \mathbb{I}[F_{var}]$ ;
    $F_{false}$ = REORDER ($F_{false}$) $\otimes \mathbb{I}[\neg F_{var}]$;
    $F_r = F_{true} \oplus F_{false}$;
    insert $F \rightarrow F_r$ in Cache;
  **return** $F_r$;
**end**

cision nodes (which occurs at the leaf when two leaf functions are compared) and these decision nodes may be out of order w.r.t. the diagram, reordering as defined in Algorithm 1 must also be applied after maximization.

On a final note, we mention that an implementation of case statements without any attempt to merge and simplify cases often cannot get past the first or second iteration of SDP; as our results show next, XADDs allow SDP to perform quite well in practice.

## 5 Empirical Results

We implemented two versions of our proposed SDP algorithm using XADDs — one that does not prune nodes of the XADD and another that uses a linear programming solver to prune unreachable nodes (for problems with linear XADDs) — and we tested these algorithms on KNAPSACK and two versions of the Mars Rover domain (adapted from [6]) that we call MARS ROVER LINEAR and MARS ROVER NONLINEAR.[4]

### 5.1 Domains

In a general MARS ROVER domain, a rover is supposed to approach one or more target points and take images of these points. Actions may consume time and energy. There are also some domain constraints, e.g., some pictures can be taken only in a certain time window and can require different levels of energy to be performed. Next we describe the two domain variants we use.

[4]While space limitations prevent a self-contained description of all domains, we note that all Java source code and a human/machine readable file format for all domains needed to reproduce the results in this paper can be found online at `http://code.google.com/p/xadd-inference`.

**MARS ROVER LINEAR** This version has two continuous variables, time $t$ and energy $e$. For each target point $i$ ($i = 1 \ldots k$), there is a boolean variable $p_i$ indicating whether the rover is at point $i$.

There are $k(k-1)$ actions $move_i$ that move the rover from point $i$ to point $j \neq i$. There are another $k$ actions $take\text{-}pic_i$ that take a picture at point $i$, which are conditioned on linear expressions over the time and energy variables. The reward is also a function of time and energy, e.g., the reward for action $take\text{-}pic_i$ is given by:

$$R_{take\text{-}pic_i}(e,t,p_i) = \begin{cases} (e > 3 + 0.0002t) \wedge (t \geq 3600) & \\ \quad \wedge (t \leq 50400) \wedge p_i : & 110 \\ \text{otherwise} : & 0 \end{cases}$$

which shows that to get a reward of 110, the rover must take a picture at point $i$ between times 3600 and 50400 with a required energy reserve that increases as the day progresses.

**MARS ROVER NONLINEAR** This version has two different continuous variables — geographic coordinates $(x, y)$ — and $k$ boolean variables $h_i$ for each picture point $i$ indicating whether the rover *has already taken* a picture of point $i$. There is a single $move$ action in this domain — it simply reduces the distance from the rover to a specific point by $\frac{1}{3}$ of the current distance; for all experiments, this target point was set to $(0, 0)$. The intent of this action is to represent the fact that a rover may move progressively more slowly as it approaches a target position in order to reach the position with high accuracy. $take\text{-}pic_i$ actions are the same from MARS ROVER LINEAR domain but conditioned by *nonlinear* expressions over the continuous $x$ and $y$ variables. The reward is also a function of $x$ and $y$, e.g., the reward for action $take\text{-}pic_i$ is given by:

$$R_{take\text{-}pic_i}(x,y,h_i) = \begin{cases} x^2 + y^2 < 4 \wedge h_i : & 0 \\ x^2 + y^2 < 4 \wedge \neg h_i : & 4 - x^2 - y^2 \\ x^2 + y^2 \geq 4 : & 0 \end{cases} \tag{12}$$

which indicates that if the rover has not already taken a picture of point $i$ and the rover is within a radius of 2 from the picture point $(0, 0)$, then the rover receives a reward that is quadratically proportional to the distance from the picture point. Hence for various points, the rover has to trade-off whether to take each picture at its current position, or to get a larger reward by first moving and potentially getting closer before taking the picture.

### 5.2 Results

For the MARS ROVER domains, we have run experiments to evaluate our SDP solution in terms of time and space cost while varying the horizon and problem size.

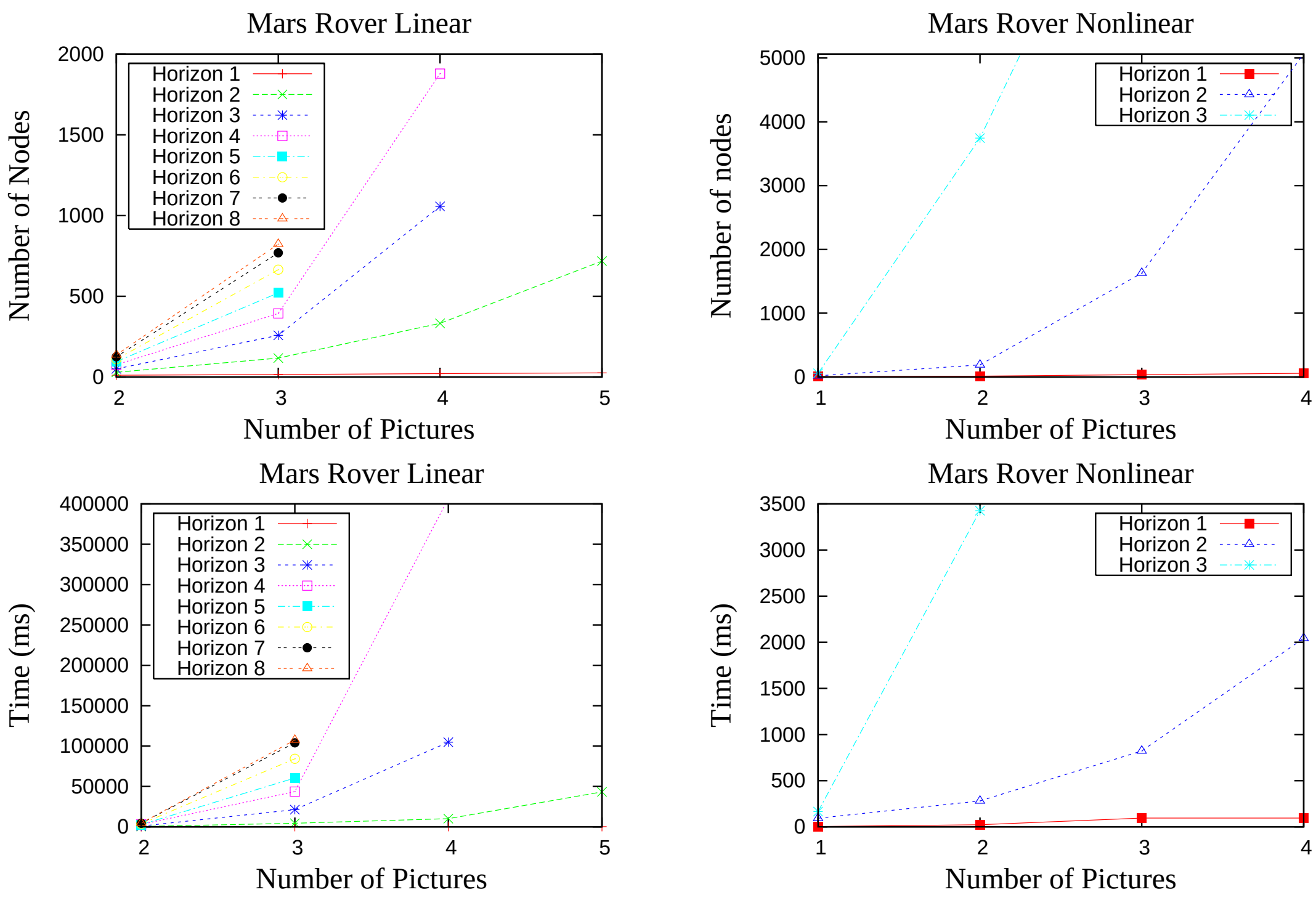

Figure 2: Space (# XADD nodes in value function) and time to optimally solve different problem sizes of the two MARS ROVER domains for varying horizon lengths.

Because the reward and transition functions for MARS ROVER LINEAR use piecewise linear case statements, we note the optimal value function in this domain is also piecewise linear. Hence in this domain, we use a linear constraint feasibility checker to prune unreachable paths in the XADD — later we will compare solutions for MARS ROVER LINEAR with and without this pruning.

In Figure 2, for both the MARS ROVER LINEAR and MARS ROVER NONLINEAR domains, we show how the number of nodes of the value function XADD (proportional to the space required to represent the value function) varies for each iteration (horizon) and different problem sizes (given by the number of pictures). We first note that the nonlinear variant appears *much* harder for SDP (much more time required and larger value functions) than for the linear variant — this is largely due to the fact that the XADD can be optimally pruned in the linear variant. Secondly, we note an apparent superlinear growth in space and time required to solve each problem as a function of the number of picture points — this likely reflects the superlinear growth of combinations of pictures that must be jointly considered as the number of pictures increases. Finally, from these graphs it is hard to summarize general algorithm behavior as a function of horizon, but it appears for the linear problem variant that both the time and space grow linearly as a function of horizon — this will be confirmed in the next experiments.

Figure 3 shows the amount of time for each iteration of SDP vs. horizon for MARS ROVER LINEAR with three picture points for SDP with and without XADD pruning. Here we see an impressive reduction in time and space as a function of horizon with pruning. Without pruning, both time and space grow super-linearly with the horizon, while with pruning time and space appear to grow linearly with the horizon.

In Figure 4, we show the exact optimal value function on the vertical axis for three domains, KNAPSACK (from Section 1), MARS ROVER LINEAR and MARS ROVER NONLINEAR, as a function of two continuous state variables shown on the horizontal axes. We notice here that the piecewise boundaries for all three plots clearly demonstrate non-rectangular boundaries. In particular, the value function plot for the MARS ROVER NONLINEAR domain demonstrates nonlinear piecewise boundaries with each piece being a nonlinear function of the state — it has the shape of stacked quadratic cones with each lower cone representing the cost of first moving from points farther away from the picture being receiving the value for taking the picture within the radius limits from (12).

To the best of our knowledge, these results demonstrate the first exact analytical solutions for DC-MDPs having optimal value functions with general linear and even nonlinear piecewise boundaries.

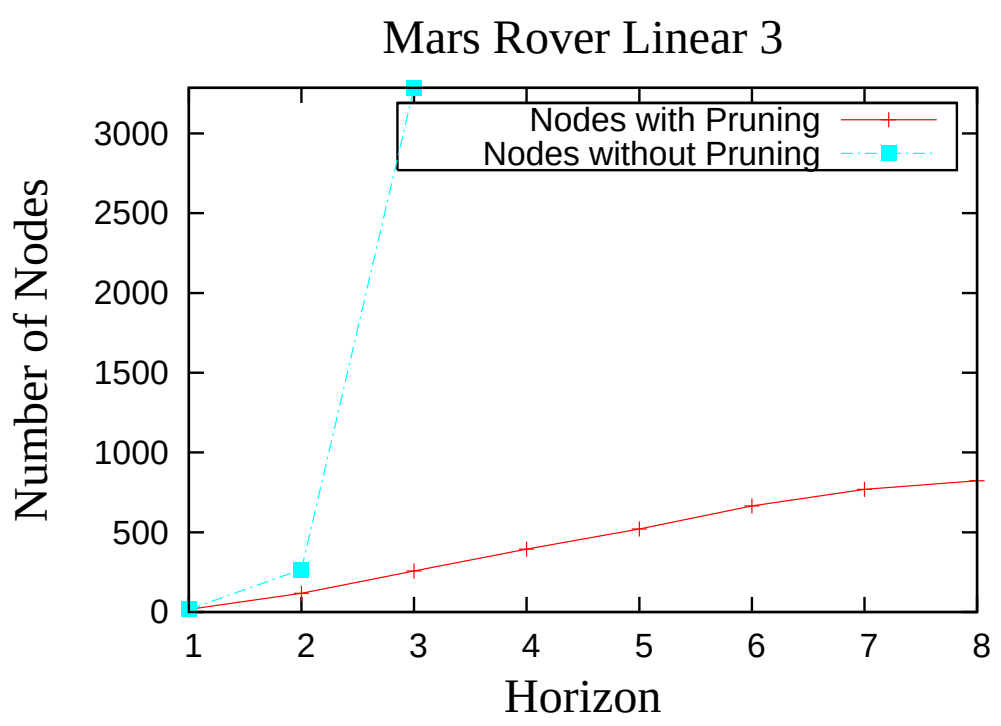

Figure 3: Space (# XADD nodes in value function) and time for different iterations (horizons) of SDP on MARS ROVER LINEAR with 3 image target points. Results shown for SDP with and without XADD infeasible path pruning.

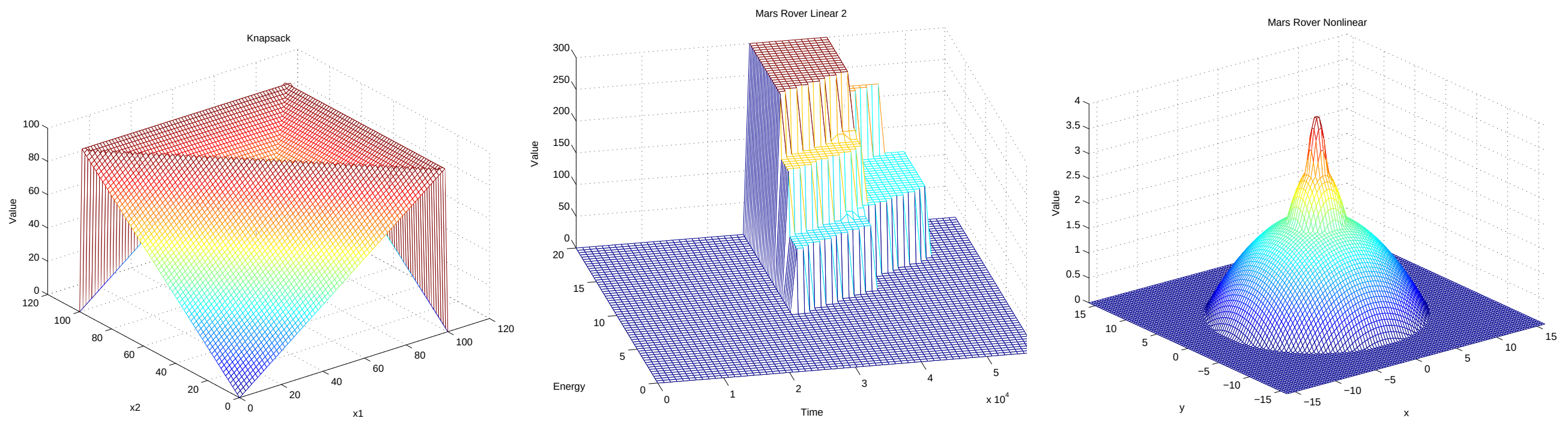

Figure 4: *Optimal* value function (vertical axis) for different domains showing *non-rectangular* piecewise boundaries. From left to right, KNAPSACK (with horizontal axes $x_2$ and $x_1$), MARS ROVER LINEAR with two pictures (with horizontal axes *energy* and *time*), and MARS ROVER NONLINEAR with one picture (with horizontal axes $y$ and $x$).

## 6 Related Work

The most relevant vein of Related work is that of [8] and [11] which can perform exact dynamic programming on DC-MDPs with rectangular piecewise linear reward and transition functions that are delta functions. While SDP can solve these same problems, it removes both the rectangularity and piecewise restrictions on the reward and value functions, while retaining exactness. Heuristic search approaches with formal guarantees like HAO* [13] are an attractive future extension of SDP; in fact HAO* currently uses the method of [8], which could be directly replaced with SDP. While [14] has considered general piecewise functions with linear boundaries (and in fact, we borrow our linear pruning approach from this paper), this work only applied to fully deterministic settings, not DC-MDPs.

Other work has analyzed limited DC-MDPS having only one continuous variable. Clearly rectangular restrictions are meaningless with only one continuous variable, so it is not surprising that more progress has been made in this restricted setting. One continuous variable can be useful for optimal solutions to time-dependent MDPs (TMDPs) [5]. Or phase transitions can be used to arbitrarily approximate one-dimensional continuous distributions leading to a bounded approximation approach for arbitrary single continuous variable DC-MDPs [12]. While this work cannot handle arbitrary stochastic noise in its continuous distribution, it does exactly solve DC-MDPs with multiple continuous dimensions.

Finally, there are a number of general DC-MDP approximation approaches that use approximate linear programming [10] or sampling in a reinforcement learning style approach [15]. In general, while approximation methods are quite promising in practice for DC-MDPS, the objective of this paper was to push the boundaries of *exact* solutions; however, in some sense, we believe that more expressive exact solutions may also inform better approximations, e.g., by allowing the use of data structures with non-rectangular piecewise partitions that allow higher fidelity approximations.

## 7 Conclusions

In this paper, we introduced a conditional stochastic equation model for the continuous part of the transition function in DC-MDPs. This representation facilitated the use of symbolic dynamic programming techniques to generate exact solutions to DC-MDPs with arbitrary reward functions and expressive nonlinear transition functions that far exceeds the exact solutions possible with existing DC-MDP solvers. In an effort to make SDP practical, we also introduced the novel XADD data structure for representing arbitrary piecewise symbolic value functions and we addressed the complications that SDP induces for XADDs, such as the need for reordering the decision nodes after some operations. All of these are substantial contributions that have contributed to a new level of expressiveness for DC-MDPS that can be exactly solved.

There are a number of avenues for future research. First off, it is important examine what generalizations of the transition function used in this work would still permit closed-form exact solutions. In terms of better scalability, one avenue would explore the use of initial state focused heuristic search-based value iteration like HAO* [13] that can be readily adapted to use SDP. Another avenue of research would be to adapt the lazy approximation approach of [11] to approximate DC-MDP value functions as piecewise linear XADDs with linear boundaries that may allow for better approximations than current representations that rely on rectangular piecewise functions. Along the same lines, ideas from APRICODD [16] for bounded approximation of discrete ADD value functions by merging leaves could be generalized to XADDs. Altogether the advances made by this work open up a number of potential novel research paths that we believe may help make rapid progress in the field of decision-theoretic planning with discrete and continuous state.

## Acknowledgements

We thank the anonymous reviewers for their comments that have helped improve the paper. The first author is supported by NICTA; NICTA is funded by the Australian Government as represented by the Department of Broadband, Communications and the Digital Economy and the Australian Research Council through the ICT Centre of Excellence program. This work has also been supported by the Brazilian agencies FAPESP (under grant 2008/03995-5) and CAPES.